\documentclass[runningheads]{llncs}
\usepackage{amsfonts}
\usepackage{bbm}
\usepackage{mathtools} % 数式表示用パッケージ
\usepackage{subcaption}
\usepackage{svg}

\usepackage{times}
\usepackage{epsfig}
\usepackage{graphicx}
\usepackage{amsmath}
\usepackage{amssymb}
\usepackage{booktabs}
\usepackage{color}
\usepackage{comment}
\usepackage{colortbl}
\usepackage{array}
\usepackage{xcolor}
\definecolor{Gray}{gray}{0.9}
\usepackage{arydshln}
\usepackage{multirow}
\usepackage{here}
\usepackage[normalem]{ulem}
\usepackage{bm} 
\useunder{\uline}{\ul}{}

\renewcommand{\paragraph}[1]{\noindent\textbf{#1}\quad}

\begin{document}
\title{Adaptive Pseudo Label Selection for Individual Unlabeled Data by Positive and Unlabeled Learning}
\titlerunning{Adaptive Pseudo Label Selection for Individual Unlabeled Data}

\author{
Takehiro Yamane\inst{1} \and
Itaru Tsuge\inst{2} \and 
Susumu Saito\inst{2} \and
Ryoma Bise\inst{1}
}
\institute{
Kyushu University, Fukuoka, Japan \and
Kyoto University Hospital, Kyoto, Japan
\\ \email{bise@ait.kyushu-u.ac.jp}
}
\authorrunning{T. Yamane et al.}

\maketitle              % typeset the header of the contribution
\begin{abstract}
This paper proposes a novel pseudo-labeling method for medical image segmentation that can perform learning on ``individual images'' to select effective pseudo-labels. We introduce Positive and Unlabeled Learning (PU learning), which uses only positive and unlabeled data for binary classification problems, to obtain the appropriate metric for discriminating foreground and background regions on each unlabeled image. Our PU learning makes us easy to select pseudo-labels for various background regions. The experimental results show the effectiveness of our method.

\end{abstract}
%
%

%%%%%%%%% BODY TEXT
% ========================================================
\section{Introduction\label{sec:intro}}
% ========================================================

Medical image segmentation plays an important role in the medical industry.
Therefore, research on automatic analysis of medical images using deep learning and artificial intelligence has become a major trend~\cite{bauer2013survey,liu2020deep,zhao2013overview}.
However, the methods of these studies require large amounts of high-quality supervised data (labeled data).

On the other hand, the cost of annotating medical images is high, and preparing a sufficient amount of labeled data is challenging.
In contrast, medical image data are abundant, and a large amount of raw data is stored in hospitals and other facilities.
Therefore, there is a need to improve the performance of deep learning models by utilizing such unannotated data (unlabeled data).

Semi-supervised learning (SSL) is a method for utilizing unlabeled data in deep learning.
In SSL, a discriminative model is trained on a small amount of supervised data and a large amount of unlabeled data.
Pseudo-labeling~\cite{lee2013pseudo,kikkawa2019ssl,ouali2020overview,van2020survey} methods, which are one of the common approaches in SSL, initially train a segmentation model using a few supervised data and select pseudo-labels based on the confidence estimated by the pre-trained model. 
Subsequently, the methods re-train the model using supervised data and the obtained pseudo labels.

\begin{figure}[t]
    \centering
    \includegraphics[width=0.92\linewidth]{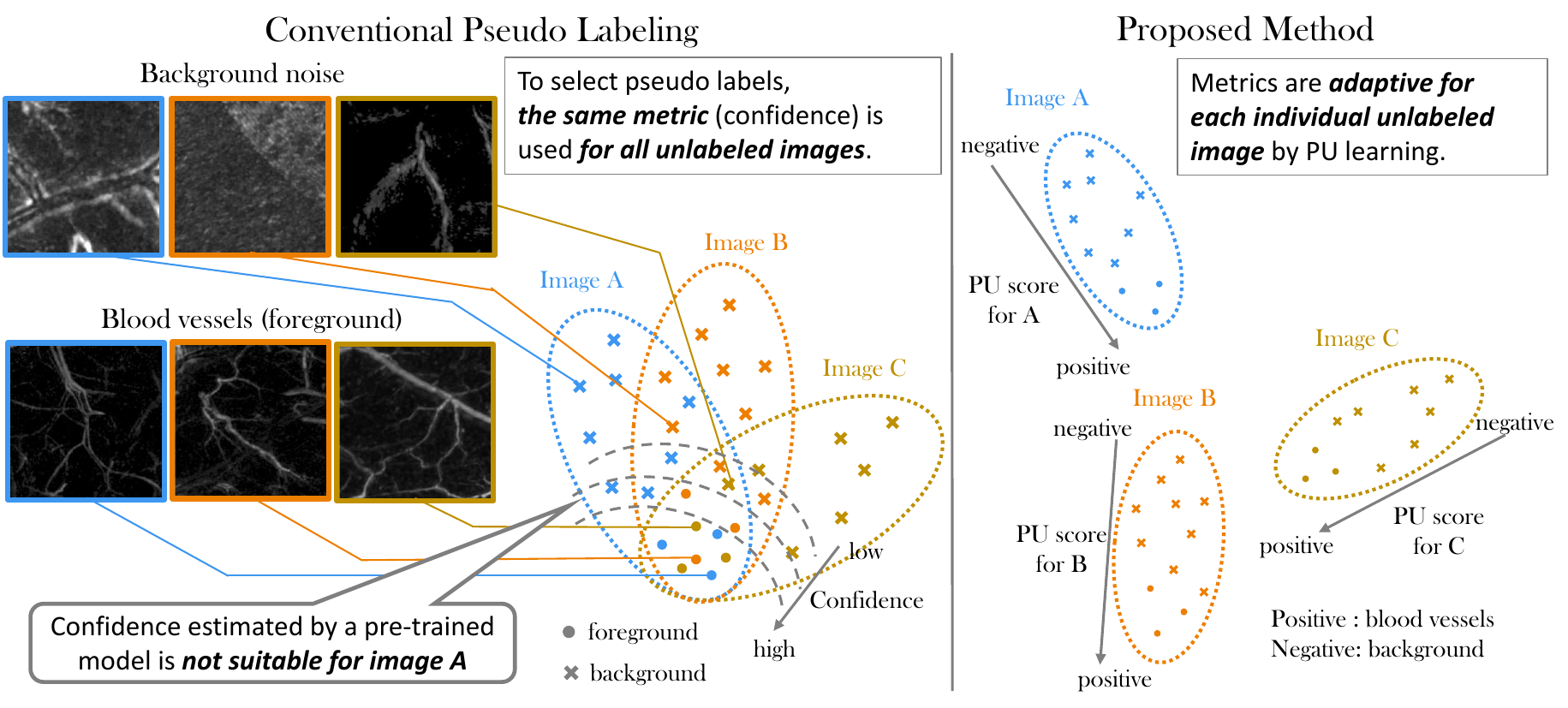}
    %\vspace{-2mm}
    \caption{Left: Overview of conventional pseudo labeling, which selects pseudo labels based on the same metric for all unlabeled images. However, the metric may not be suitable for a specific image. Right: Overview of the proposed method, which adaptively defines a metric for each individual image using PU learning.}
    \label{fig:idea}
    %\vspace{-3mm}
\end{figure}

Conventional pseudo-labeling methods use the same metric (confidence) for all unlabeled images, as shown in Fig.~\ref {fig:idea} (Left). 
However, the common metric for all unlabeled images may not be appropriate for a specific image. This is because medical images contain various types of background noise, such as reconstruction artifacts, sensor noises, body motions, etc. This indicates that each image's feature distribution is biased, as shown in Fig.~\ref {fig:idea} (Left). Determining the confidence threshold under a common metric becomes challenging; incorrect labeling due to an unsuitable threshold harms learning. Specifically, the low threshold makes pseudo-labels containing noisy labels. Moreover, setting the threshold too high to maintain high accuracy in pseudo-labeling makes obtaining a sufficient number of pseudo-labels difficult.

Therefore, we propose a novel SSL method that leverages the feature distribution of each unlabeled image to select appropriate pseudo-labels, as shown in Fig.~\ref{fig:idea} (Right).
We employ Positive and Unlabeled Learning (PU learning) to select pseudo-labels for each unlabeled image. PU learning is a binary classification learning method designed for situations where only a subset of labels is positive while others remain unlabeled.
It has been used in several studies that deal with data sets with incomplete labeling~\cite{fujii2021cell,KikkawaR2021,zhao2021positive}.
Our method first extracts reliable pseudo-labels using a conventional pseudo-label method. 
Next, using PU learning, the method selects additional pseudo-labels from the remaining unlabeled pixels for individual unlabeled images.
Finally, re-training is performed using a small number of labeled data and the obtained pseudo-labels.
Obtaining an appropriate metric for each individual unlabeled image to select pseudo-labels enables the acquisition of more pseudo-labels and improves accuracy.

In the experiments, we conduct segmentation experiments using two public datasets of fundus images and experiments on estimating vascular structure in 3D photoacoustic images, which are clinical data.
Three-dimensional photoacoustic images~\cite{li2009photoacoustic} are expected to be used for mapping blood vessels before surgical procedures.
Through these experiments, we demonstrate the effectiveness of this method and its potential for application.

In summary, the main contributions of this paper are as follows.
\begin{itemize}
\item We propose a semi-supervised method that can obtain high-quality pseudo-labels by leveraging the feature distribution of each unlabeled image using PU learning. 
\item The effectiveness of the proposed method for two applications is demonstrated using two public datasets and one clinical data.
\end{itemize}

% ========================================================
%\section{Related work\label{sec:relatedwork}}
% ========================================================

\begin{figure}[t]
    \centering
    \includegraphics[width=0.9\linewidth]{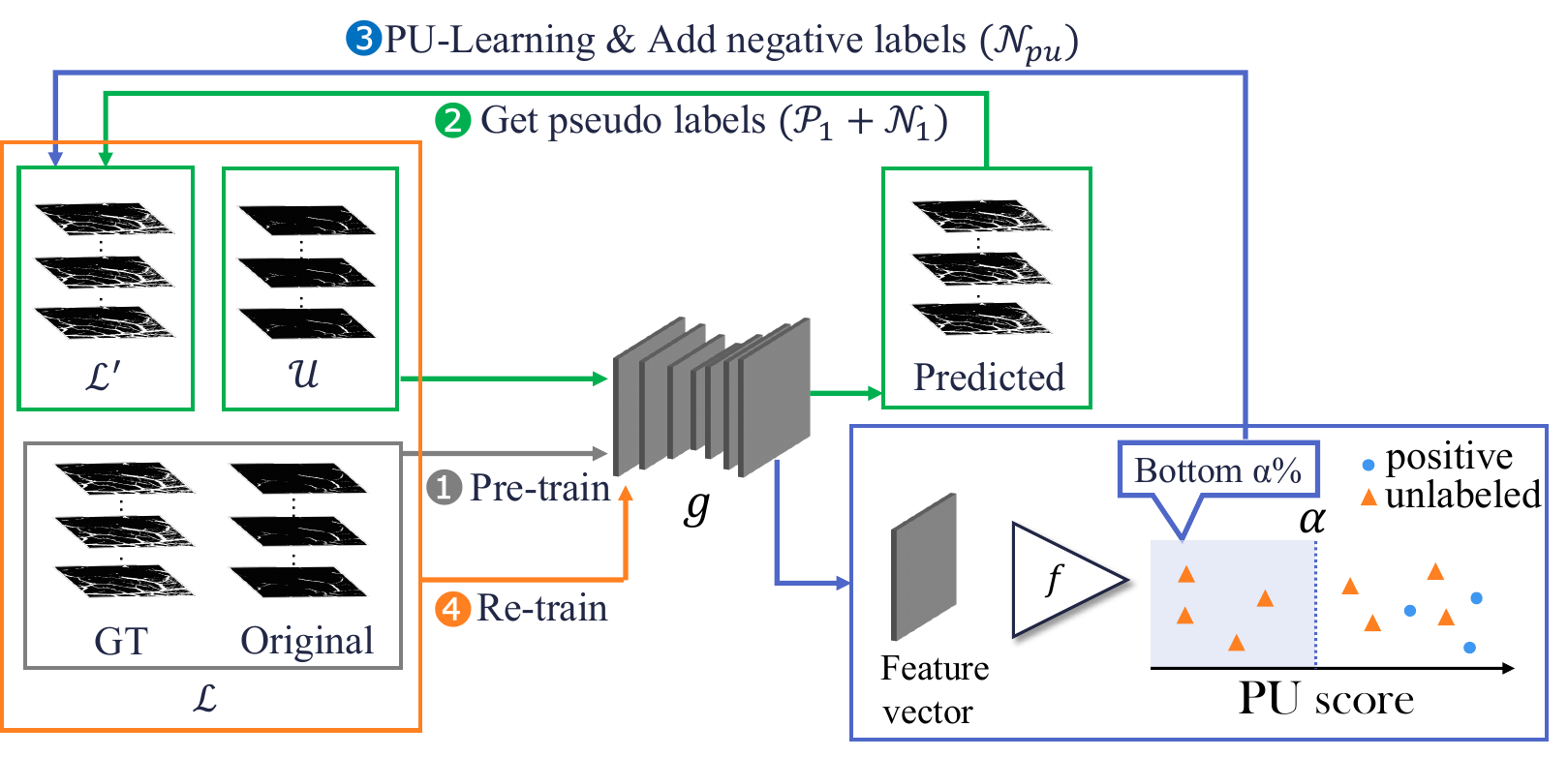}
    \caption{Overview of the proposed method, which has four steps; 1) pre-training by supervised data; 2) pseudo-labeling by the confidence estimated by the pre-trained model; 3) additional pseudo-labeling based on the adaptive learning for each individual image; 4) re-training by the supervised and pseudo labels.}
    \label{fig:proposed}
    \vspace{-3mm}
\end{figure}

%%%%%%%%%%%%%%%%%%%
\section{Semi-supervised learning with pseudo-label selection using PU Learning}

\noindent{\textbf{Overview:}}
% \subsection{Overview}
The objective of this task is that, given an original image $I_i$, a network $g$ produces a segmentation mask.    
In this task, a set of original images and the corresponding segmentation mask $\mathcal{L} = \{ {I_i}, {A_i} \}_{i=1}^{N_s}$ is given as supervised data, and $N_u$ images $\mathcal{U} = \{ {I_i} \}_{i=1}^{N_u}$ as unlabeled data, we train the network $g$ using $\mathcal{L}$ and $\mathcal{U}$.

Figure~\ref{fig:proposed} shows an overview of the proposed method, which consists of four steps:
1) pre-training using supervised data, which trains the network $g$ to produce the segmentation mask from the original image using the supervised data $\mathcal{L}$; 2) pseudo-labeling based on the estimated confidence, which obtains pseudo-labels for both foreground and background regions based on the pixel's confidence in the unlabeled data $\mathcal{U}$ estimated by the pre-trained model $g$; 3) pseudo-labeling for individual unlabeled images using PU learning~\cite{elkan2008learning}, which obtains the pseudo-labels for background regions; and 4) re-training using $\mathcal{L}$ and $\mathcal{U}$ with the obtained pseudo-labels.
The details of each step are described below.

%\noindent{\textbf{Supervised learning:}}
\subsection{Pre-training by supervised data}
The network $g(\cdot;\theta_{sv})$ is trained to produce the corresponding segmentation mask $A_i \in \mathcal{L}$ from the original image $I_i \in \mathcal{L}$, in which we can use any backbones, such as U-Net~\cite{Ronneberger2015} and FR-UNet~\cite{liu2022full}.
The parameter of the pre-trained network is represented as $\theta_{sv}$.
We use the pixel-level cross-entropy between the estimation and the ground truth as follows:
\begin{equation}
\label{eq:L_s}
L_s = \sum_{I_i, A_i \in \mathcal{L}}{\sum_{j \in \Omega_i}{A_i(j)\log g(I_i(j);\theta_{sv})}},
\end{equation}
where $I_i(j)$ and $A_i(j)$ are pixel values of the $i$-th pixel in $I_i$ and $A_i$, respectively.

%\noindent{\textbf{Obtaining pseudo-labels:}}
\subsection{Pseudo-labeling based on confidence}
Next, we obtain the pixel-level pseudo-labels based on the confidence estimated by the pre-trained network $g(\cdot;\theta_{sv})$.
Given an unsupervised image $I_i \in \mathcal{U}$, we first estimate the confidence map $\hat{A}_i=g(I_i;\theta_{sv})$ using the pre-trained network, where $0 \leq \hat{A}_i(j) \leq 1$, and 0 and 1 indicate the background and the foreground, respectively.
Then, the set of the image IDs and the pixel index on positive labels (vessels) $\mathcal{P}_1=\{(i,j)|\hat{A}_i(j)>th_p\}$ and those on negative labels (background) $\mathcal{N}_1=\{(i,j)|\hat{A}_i(j)<th_n\}$ are obtained as pseudo-labels in the image based on thresholds, where $j$ is the pixel-index on the $i$-th image.
The thresholds $th_p$ and $th_n$ have a trade-off with their values.
For example, if $th_p$ takes a high value, we can obtain the pseudo-labels accurately, but the number of pseudo-labels is limited; if $th_p$ takes a low value, many pseudo-labels can be obtained. However, the pseudo-labels may contain noises.
Since the background regions contain many noises, this method uses strict thresholds (the lower threshold for $th_n$) for pseudo-labels of the background region to obtain accurate pseudo-labels.

%\noindent{\textbf{Adding pseudo-labels using PU Learning:}}
\subsection{Adaptive pseudo-labeling for each image using PU learning}
In this step, the method tries to obtain additional pseudo-labels for background regions based on the individual feature distribution of each unlabeled image.
To use the feature distribution of each unsupervised image for pseudo-labeling, we introduce PU learning~\cite{elkan2008learning}, which can train a network using positive and unlabeled data with representation learning.

We use the feature vectors of the foreground regions in the supervised images as the positive labels $\mathcal{X}_p=\{\bm{x}_i(p)\}_{(i,p) \in \mathcal{L}_p}$, where $\mathcal{L}_p$ is the set of the image and pixel indices that have a positive value; $\mathcal{L}_p=\{(i,j)|A_i(j)=1\}$,
$\bm{x}_i(p)$ indicates the feature vector at the position $p$ corresponding to the image $I_i$, and it is extracted by the feature extraction layers of the network $g$ (i.e., the output vector before the final output layer), denoted as $g'$.

The set of feature vectors $\mathcal{X}_u^{(i)}=\{\bm{x}_i(u)\}_{u \in \mathcal{U}_i}$ on pixels except the pseudo-labels are used as unlabeled data, where $\mathcal{U}_i = \Omega^{(i)} - (\mathcal{P}_1^{(i)} + \mathcal{N}_1^{(i)})$. $\mathcal{P}_1^{(i)}$ and $\mathcal{N}_1^{(i)}$ are the pixel indices of the positive and negative pseudo-labels in image $I_i$, respectively, and $\Omega^{(i)}$ is a set of all the pixel indices in the image $I_i$. Note that $\mathcal{U}_i$ only contains ambiguous pixels since the high-confidence pixels are already extracted in the previous step.
Using PU learning, we train a binary classification function $f(\cdot;\theta_{pu}^{(i)})$, implemented as a fully connected layer, with fixed feature extractor $g'(\cdot;\theta_{sv})$ trained by supervised learning. The PU loss $L_{pu}$~\cite{NIPS2017_7cce53cf} is defined as:
\footnotesize
\begin{eqnarray}
\hspace{-8mm}&& L_{pu}=\pi_p\hat{R}_{p}^{+}(f) + \max\{0,\hat{R}_{u}^{-}(f)-\pi_p \hat{R}_{p}^{-} (f) \},\notag \\
\hspace{-8mm}&& \hat{R}_{p}^{+} (f) = (1/n_p)\sum_{i=1}^{n_p}l(f(\bm{x}_i^p), +1),
\hspace{3mm} \hat{R}_{u}^{-} (f) = (1/n_u)\sum_{i=1}^{n_u}l(f(\bm{x}_i^u), -1),\notag \\
\hspace{-8mm}&&\hat{R}_{p}^{-} (f) = (1/n_p)\sum_{i=1}^{n_p}l(f(\bm{x}_i^p), -1),
\end{eqnarray}
\normalsize
where $n_p$ is the total number of $\mathcal{X}_p^{(i)}$, $n_u$ is the total number of $\mathcal{X}_u^{(i)}$, $x_i^p\in\mathcal{X}_p^{(i)}$ and $x_i^u\in\mathcal{X}_u^{(i)}$, and $l$ is the zero-one loss. The positive ratio $\pi_p$ is estimated using labeled data: $\pi_p=\frac{n_p}{n_p+n_u}$.

Then, we select additional pseudo labels for individual unlabeled image $I_i \in \mathcal{U}$ using $f(\cdot;\theta_{pu}^{(i)})$.
The pixel whose output value $f(\bm{x}_i(u);\theta_{pu}^{(i)})$ (PU-score: likelihood for the positive label) is in the lower $\alpha \%$ is obtained as a negative label, and it is added into $\mathcal{N}_{pu}$.

This process for pseudo-label selection by PU learning is iteratively conducted for every individual image and the set of pseudo-labels $\mathcal{N}_{pu}$ is finally obtained.
Since PU learning is applied for each unlabeled image $I_i \in \mathcal{U}$, the method can use the feature distribution of each image to select pseudo-labels. Therefore, pseudo-labels can cover the variations of the noises (negative samples) in each image.

%\noindent{\textbf{Re-training with pseudo-labels:}}
\subsection{Re-training with pseudo-labels}
The network $g$ is re-trained using the supervised data $\mathcal{L}$ and the pseudo-labels $\mathcal{P}_1$, $\mathcal{N}_1$, and $\mathcal{N}_{pu}$ obtained by the above procedure.
The loss function is defined as follows:
\footnotesize
\begin{equation}
Loss = \sum_{\{I_i(j), A_i(j)\} \in \mathcal{L}}{L(g(I_i(j)),A_i(j))}
+ \sum_{I_i(j) \in \mathcal{L}'}{L(g(I_i(j)),P_i(j))},
\end{equation}
\normalsize
where $\mathcal{L}' = \mathcal{P}_1 \cup \mathcal{N}_1 \cup \mathcal{N}_{pu}$, 
$P_i(j)$ is a pseudo-label (positive (+1) or negative (0)), and $L$ is a loss function, which takes an entropy loss.
Note that the loss is only calculated for pixels contained in the pseudo labels.
In inference, an estimation result is obtained by inputting images into the re-trained model.

% ========================================================
\section{Experiments}
% ========================================================

% --------------------------------------------------------
\subsection{Vessel segmentation on public datasets}
\noindent{\textbf{Experimental setup:}}
To show the effectiveness of segmentation tasks, we evaluated our method using two public datasets: DRIVE~\cite{DRIVE:2014} and CHASE DB~\cite{ChaseDB:2020}.
In the experiment with DRIVE, we used two images as supervised data $\mathcal{L}$, four images as unlabeled data $\mathcal{U}$, and the rest of the images as test data in each cross-validation to mimic the situation when the number of labeled data is few.
With CHASE DB, two images for $\mathcal{L}$, five images for $\mathcal{U}$, and the rest for test data were used.
Dice was used as the evaluation metric, and the average of Dice was evaluated in 5-fold cross-validation.

In training using supervised and pseudo-labels, the data were augmented with random crop, random rotation, and flip. The network was trained over 2000 epochs using Adam optimizer.
For pseudo-label acquisition, $th_p$ was set to 0.8 and $th_n$ to 0.1, and for pseudo-label selection by PU Learning, $\alpha=20\%$.

\vspace{0.5\baselineskip}
\noindent{\textbf{Evaluation:}}
We evaluated four methods as comparative methods: 1) Baseline that trains the segmentation network by supervised learning; 2) MNS~\cite{lou2023min} that introduces contrastive learning for medical image segmentation, which is one of the state-of-the-art methods; 3) Pseudo (w/o pu), which is conventional pseudo-labeling, in which we modified the method~\cite{lee2013pseudo} that is originally designed for classification to segmentation; 4) Ours (batch), which additionally selects the pseudo-labels with applying PU-learning to all unlabeled data at once not for individually; and 5) Ours, which additionally selects the pseudo-labels with applying PU-learning to individual unlabeled data.

In addition, to show the effectiveness of our method for different backbones, we applied the above methods to two different backbones: U-Net~\cite{Ronneberger2015}, which has been widely used in segmentation tasks, and FR-UNet~\cite{liu2022full}, which is a state-of-the-art method for blood vessel segmentation (the best method on the leaderboard of DRIVE~\cite{pwc}). 
As reference values, we also prepared the results of supervised learning (fully supervised) for DRIVE and CHASE DB, which use all data except test data as supervised data.

\begin{table*}[t]
 \caption{Evaluation results for each dataset by Dice using two datasets (DRIVE and CHASE DB). ``pl'' shows pseudo-labels obtained by thresholding. ``pu'' shows negative labels added by PU Learning. ``ind'' shows that PU learning is applied for individual unlabeled image.}
 \label{tab:q-eval_open}
 \centering
 \begin{tabular}{l|ccc|p{18mm}:p{18mm}|p{18mm}:p{18mm}|c}
   \hline
    \multirow{2}{*}{Method} & \multirow{2}{*}{pl} &\multirow{2}{*}{pu} &\multirow{2}{*}{ind} & \multicolumn{2}{c|}{Backbone : U-Net} & \multicolumn{2}{c|}{Backbone : FR-UNet} & \multirow{2}{*}{Avg.}\\ \cline{5-8}
    & & & & DRIVE & CHASE & DRIVE & CHASE\\
   \hline\hline 
    Baseline & & & & 0.787  & 0.730 & 0.784  & 0.801 & 0.776\\
    \hline
    MMS~\cite{lou2023min} & & & & 0.733  & 0.699 & 0.740  & 0.701 & 0.718\\
    \hline
    Pseudo w/o pu & \checkmark & & & 0.795 & 0.809 & 0.765 & 0.774 & 0.786\\
    \hline
    Ours (batch)  & \checkmark & \checkmark & & 0.720 & 0.737 & \textbf{0.789} & 0.791 & 0.759\\
   \hline
    Ours & \checkmark & \checkmark & \checkmark & \textbf{0.796} & \textbf{0.810} & 0.786 & \textbf{0.804} & \textbf{0.799}\\
   \hline\hline
    Fully supervised & & & & 0.805 & 0.823 & 0.807 & 0.835 & 0.818\\
   \hline
  \end{tabular}
  %\vspace{-2mm}
\end{table*}

Table~\ref{tab:q-eval_open} shows the results of the comparison of segmentation performance.
When the training data is insufficient, the performances of U-Net and FR-UNet decrease.
MMS's performance was worse than Baseline.
This method prepares the negative pairs from different locations in unlabeled images for contrastive learning. We consider that both the paired images of negatives may contain blood vessels, adversely affecting the performance.
Conventional pseudo-labeling increased the performance on the U-net backbone but decreased the performance on the FR-UNet backbone. It shows that confidence-based pseudo-labeling risks containing noisy pseudo-labels and reducing performance.
Ours (PU batch) decreased the performance. This indicates that applying PU learning to all unlabeled images is difficult due to the variation of background noises.
In contrast, our method was the best in any dataset and backbone and outperformed the comparative methods on average.
Furthermore, the fact that the results of Ours are better than those of Ours (batch) confirms the effectiveness of adapting PU Learning to individual images.

\begin{figure}[t]
 \centering
 \includegraphics[width=\linewidth]{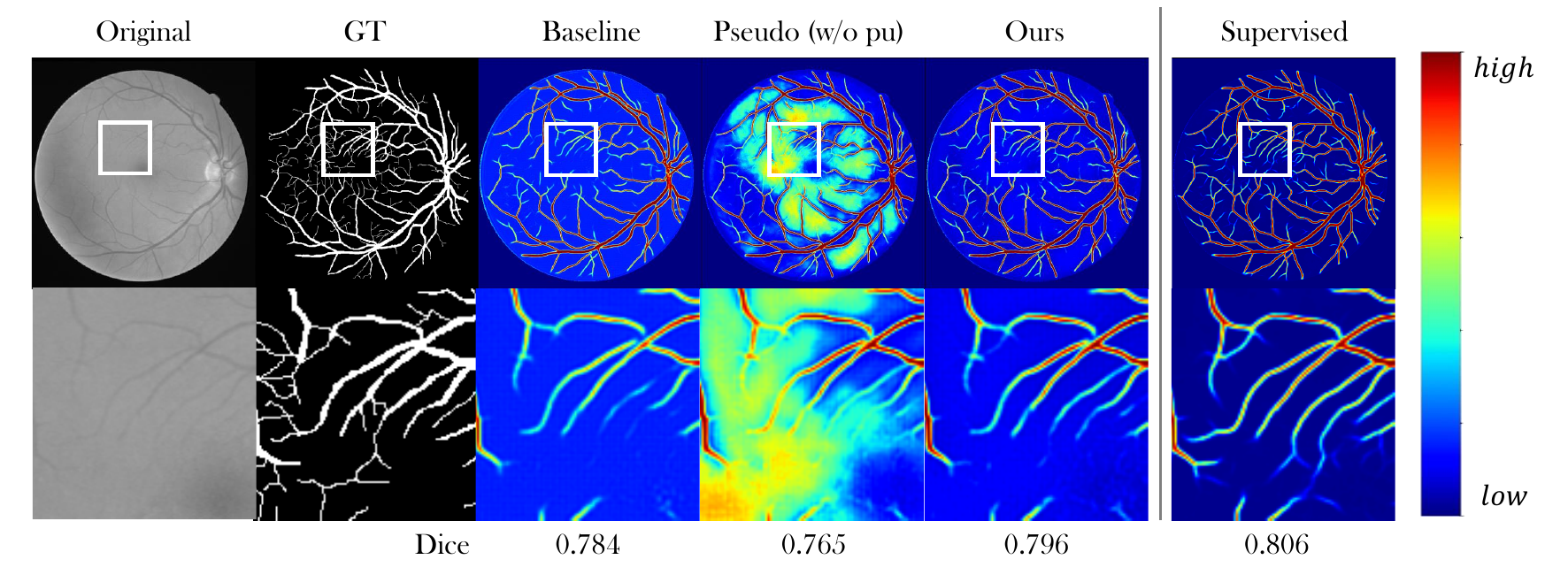}
  %\vspace{-5mm}
  \caption{Original image, ground truth, Baseline estimation results, Pseudo (without PU), Our method, and supervised learning results from left to right. Top: the entire images, Bottom: the enlarged images.}\label{result}
  % \vspace{-1zh}
  % \vspace{-3mm}
\end{figure}

Fig.~\ref{result} shows examples of the estimation results by each method. In the Baseline results, some vessels were not clearly extracted. In Pseudo (w/o pu) results, the confidence on background regions also takes higher values due to inaccurate pseudo labels (false positives). Our method improved vessel extraction, successfully capturing vessels that were unclear in the Baseline result. Although the results are similar to the supervised results, thin and unclear vessels were not consistently extracted across all methods due to the dataset's volume limitations in labeled and unlabeled images. An analysis of the effectiveness of the number of unlabeled images is needed in future work.

\subsection{Vascular structure estimation of 3D photoacoustic images in clinical data}
To demonstrate the effectiveness of various tasks, we additionally conducted experiments using Photoacoustic (PA) images~\cite{li2009photoacoustic,bise2016vascular}, which are used in a real clinical application. The application is operation planning in flap surgery, where the PA images are used to check the vascular map during surgery. Currently, the vessels are manually traced in 3D by medical technicians, and it is used to check the vascular map during surgery. In this experiment, we applied our method to obtain the vascular structures automatically.

3D photoacoustic (PA) images were captured at Kyoto University Hospital, which was used for planning flap surgery.
Making segmentation masks of complex structures in 3D PA images is complicated.
Instead of segmentation masks, we traced the center line of the blood vessels as annotations, which were recorded as a set of coordinates around the center.
We generate a clear vessel image (called vessel heatmap) as the ground truth $A_i$ by adding Gaussian blur to the annotated center points, where we followed the cell position heatmap for cell detection~\cite{nishimura2019weakly}.

Despite the significantly low contrast of vessels in deep regions, their signals exhibit high contrast in the ground-truth heatmap. Using the ground truth of the vessel heatmap, the network is trained to estimate the clear image that the low-visible vessels become clear. It makes vessel tracking easier. After obtaining the heatmap, we applied a simple vessel tracing method to obtain the vessel structure, similar to \cite{BiseR2016}.

\vspace{0.5\baselineskip}
\noindent{\textbf{Experimental setup:}}
We used U-Net or FR-UNet as the network's backbone $g$ in this task. We performed data augmentation, e.g., random crop, rotation, and flip, and used the Adam optimizer. The number of epochs was set to 1000.
For pseudo-labeling, $th_p$ was set to 100/256 and $th_n$ to 2/256. In PU learning, the network was trained with 2000 epochs using Adam. We set $\alpha=20\%$ for pseudo-label selection after PU learning.

\vspace{0.5\baselineskip}
\noindent{\textbf{Evaluation:}}
To show the proposed method for vascular structure estimation, we evaluated the accuracy of vessel tracing with comparative methods: 1) supervised learning; 2) Pseudo (w/o pu), which retrained the network using pseudo-labels without using PU learning; and 3) our method (Ours(PU)).
These three methods used two backbone networks (U-Net and FR-Unet). 
The same tracing algorithm was applied after estimating the vessel heatmaps for all methods.

\begin{table}[t]
 \caption{Evaluation results for 3D photoacoustic images.}
 \label{q-eval}
 \centering
 \begin{tabular}{l|c|p{8mm}:p{8mm}:p{8mm}:p{8mm}}
   \hline
   Backbone & Method & 3px & 6px & 9px & Avg.\\
   \hline \hline
   \multirow{3}{*}{U-Net} & Baseline &  0.635 & 0.725 & 0.775 & 0.712\\
   %\hline
   \cline{2-6}
   & Pseudo & 0.528 & 0.679 & 0.743 & 0.650\\
   \cline{2-6}
   & Ours & \textbf{0.647} & \textbf{0.793} & \textbf{0.849} & \textbf{0.763}\\
   \hline
   \hline 
   \multirow{3}{*}{FR-UNet} & Baseline & 0.832 & 0.884 & 0.913 & 0.876\\
   %\hline
   \cline{2-6}
   & Pseudo & 0.863 & 0.924 & 0.949 & 0.912\\
   \cline{2-6}
    & Ours & \textbf{0.871} & \textbf{0.928} & \textbf{0.953} & \textbf{0.917}\\
   \hline
  \end{tabular}
  %\vspace{-3mm}
\end{table}

The performance metric is defined as the coverage of the annotation given near the vessel's center, which was designed based on \cite{schaap2009standardized}. 
The coverage is calculated by the error between the annotation and tracing results over a tolerance (threshold) for each blood vessel branch, and the average coverage of them is evaluated. 
This metric has a hyper-parameter about the threshold to determine the true positive traces. Thus, we evaluated the method while changing the threshold (3, 6, and 9 pixels) and calculated their average.

Table~\ref{q-eval} shows the vessel tracing performance of each method.
Pseudo (w/o pu) improved the performance on FR-UNet but decreased it on U-Net. It shows that pseudo-label selection is sensitive and sometimes has a negative effect.
Our method improved the tracing performance significantly from backbone methods (U-Net and FR-UNet) by adding negative labels (background) based on PU learning.
As a result, our method outperformed the baseline methods in all metrics.
Planning flap surgery aims to enable medical doctors to recognize the vessel structures. Therefore, a 6-9 pixels shift from the annotated center is sufficient to check the structures. In particular, vessel structures in the deep area, which are the most important vessels in the surgery, can be clearly extracted using our method.

% ========================================================
\section{Conclusion}
% ========================================================
This paper proposes a semi-supervised learning for medical image segmentation that can adaptively select pseudo-labels in ``individual'' unlabeled images.
The method can perform learning on ``individual images” to select effective pseudo-labels by PU learning.
In experiments, segmentation experiments were conducted on retinal vessel segmentation, and the segmentation performance was improved using pseudo-labels obtained by the proposed method.
In addition, as an application to real clinical data, we performed blood vessel structure estimation in 3D photoacoustic images and showed the effectiveness of our method for real data.
Our future work includes expanding the proposed method to other segmentation tasks, such as organ segmentation. We also plan to refine the selection of pseudo-labels for foreground regions.
% ========================================================

\vspace{0.5\baselineskip}
{\noindent{\bf Acknowledgement}:
This work was supported by JSPS-JP23KJ1723, SIP-JPJ01242 and AMED JP19he2302002.}

% \clearpage

% \bibliographystyle{splncs04}
% \bibliography{main}
\bibliographystyle{splncs04}
\bibliography{main}

\begin{thebibliography}{10}
\providecommand{\url}[1]{\texttt{#1}}
\providecommand{\urlprefix}{URL }
\providecommand{\doi}[1]{https://doi.org/#1}

\bibitem{bauer2013survey}
Bauer, S., et~al.: A survey of mri-based medical image analysis for brain tumor studies. Physics in Medicine \& Biology  \textbf{58}(13), ~R97 (2013)

\bibitem{BiseR2016}
Bise, R., et~al.: 3d structure modeling of dense capillaries by multi-objects tracking. In: IEEE Conference on Computer Vision and Pattern Recognition Workshops (CVMI). pp. 29--37 (2016)

\bibitem{bise2016vascular}
Bise, R.e.a.: Vascular registration in photoacoustic imaging by low-rank alignment via foreground, background and complement decomposition. In: MICCAI. pp. 326--334 (2016)

\bibitem{elkan2008learning}
Elkan, C., et~al.: Learning classifiers from only positive and unlabeled data. In: Proceedings of the 14th ACM SIGKDD international conference on Knowledge discovery and data mining. pp. 213--220 (2008)

\bibitem{fujii2021cell}
Fujii, K., et~al.: Cell detection from imperfect annotation by pseudo label selection using p-classification. In: International Conference on Medical Image Computing and Computer-Assisted Intervention. pp. 425--434. Springer (2021)

\bibitem{kikkawa2019ssl}
Kikkawa, R., Sekiguchi, H., Tsuge, I., Saito, S., Bise, R.: Semi-supervised learning with structured knowledge for body hair detection in photoacoustic image. In: ISBI (2019)

\bibitem{KikkawaR2021}
Kikkawa, R., et~al.: Unsupervised body hair detection by positive-unlabeled learning in photoacoustic image. In: 2021 43rd Annual International Conference of the IEEE Engineering in Medicine \& Biology Society (EMBC). pp. 3349--3352 (2021)

\bibitem{NIPS2017_7cce53cf}
Kiryo, R., et~al.: Positive-unlabeled learning with non-negative risk estimator. In: Guyon, I., Luxburg, U.V., Bengio, S., Wallach, H., Fergus, R., Vishwanathan, S., Garnett, R. (eds.) Advances in Neural Information Processing Systems. vol.~30. Curran Associates, Inc. (2017)

\bibitem{lee2013pseudo}
Lee, D.H., et~al.: Pseudo-label: The simple and efficient semi-supervised learning method for deep neural networks. In: Workshop on challenges in representation learning, International Conference on Machine Learning. vol.~3, p.~896 (2013)

\bibitem{li2009photoacoustic}
Li, C., et~al.: Photoacoustic tomography and sensing in biomedicine. Physics in Medicine \& Biology  \textbf{54}(19), ~R59 (2009)

\bibitem{liu2022full}
Liu, W., et~al.: Full-resolution network and dual-threshold iteration for retinal vessel and coronary angiograph segmentation. IEEE Journal of Biomedical and Health Informatics  \textbf{26}(9),  4623--4634 (2022)

\bibitem{liu2020deep}
Liu, Y., et~al.: A deep learning system for differential diagnosis of skin diseases. Nature medicine  \textbf{26}(6),  900--908 (2020)

\bibitem{lou2023min}
Lou, A., et~al.: Min-max similarity: A contrastive semi-supervised deep learning network for surgical tools segmentation. IEEE Transactions on Medical Imaging  (2023)

\bibitem{nishimura2019weakly}
Nishimura, K., et~al.: Weakly supervised cell instance segmentation by propagating from detection response. In: International Conference on Medical Image Computing and Computer-Assisted Intervention. pp. 649--657. Springer (2019)

\bibitem{ouali2020overview}
Ouali, Y., et~al.: An overview of deep semi-supervised learning. arXiv preprint arXiv:2006.05278  (2020)

\bibitem{ChaseDB:2020}
Owen, C.G., et~al.: {Measuring Retinal Vessel Tortuosity in 10-Year-Old Children: Validation of the Computer-Assisted Image Analysis of the Retina (CAIAR) Program}. Investigative Ophthalmology \& Visual Science  \textbf{50}(5),  2004--2010 (2009)

\bibitem{pwc}
PapersWithCode: Retinal vessel segmentation on drive leaderboard (2023), \url{https://paperswithcode.com/sota/retinal-vessel-segmentation-on-drive}

\bibitem{Ronneberger2015}
Ronneberger, O., et~al.: {U-Net: Convolutional networks for biomedical image segmentation}. Lecture Notes in Computer Science (including subseries Lecture Notes in Artificial Intelligence and Lecture Notes in Bioinformatics)  \textbf{9351},  234--241 (2015)

\bibitem{schaap2009standardized}
Schaap, M., et~al.: Standardized evaluation methodology and reference database for evaluating coronary artery centerline extraction algorithms. Medical image analysis  \textbf{13}(5),  701--714 (2009)

\bibitem{DRIVE:2014}
Staal, J., et~al.: Ridge-{B}ased {V}essel {S}egmentation in {C}olor {I}mages of the {R}etina. IEEE Transactions on Medical Imaging  \textbf{23}(4),  501--509 (2004)

\bibitem{van2020survey}
Van~Engelen, J.E., Hoos, H.H.: A survey on semi-supervised learning. Machine learning  \textbf{109}(2),  373--440 (2020)

\bibitem{zhao2013overview}
Zhao, F., Xie, X.: An overview of interactive medical image segmentation. Annals of the BMVA  \textbf{2013}(7),  1--22 (2013)

\bibitem{zhao2021positive}
Zhao, Z., et~al.: Positive-unlabeled learning for cell detection in histopathology images with incomplete annotations. In: International Conference on Medical Image Computing and Computer-Assisted Intervention. pp. 509--518. Springer (2021)

\end{thebibliography}

% ========================================================

\end{document}